Cameron Morin and Matti Marttinen Larsson*

# Large corpora and large language models: a replicable method for automating grammatical annotation

**Abstract:** Much linguistic research relies on annotated datasets of features extracted from text corpora, but the rapid quantitative growth of these corpora has created practical difficulties for linguists to manually clean and annotate large data samples. In this paper, we present a method that leverages large language models for assisting the linguist in grammatical annotation through prompt engineering, training, and evaluation. We apply this methodological pipeline to the case study of formal variation in the English evaluative verb construction "*consider* X (as) (to be) Y", based on the large language model Claude 3.5 Sonnet and data from Davies's NOW and Sketch Engine's EnTenTen21 corpora. Overall, we reach a model accuracy of over 90 % on our held-out test samples with only a small amount of training data, validating the method for the annotation of very large quantities of tokens of the construction in the future. We discuss the generalizability of our results for a wider range of case studies of grammatical constructions and grammatical variation and change, underlining the value of AI copilots as tools for future linguistic research, notwithstanding some important caveats.

# 1 Introduction

Corpus linguistic research typically works with vast quantities of data, which appear to have only kept growing since the 1990s and the early 2000s in the context of the "quantitative turn" undergone by the field (Kortmann 2021). Researchers across linguistics have borne witness to a "march of data" underpinned by access to larger sources and improvements in the technological means to process data and build datasets, among other factors (Coats and Laippala 2024). For corpus-based research specifically, this means that linguists have gained access to increasingly massive corpora of natural language, in English and other languages. Notable examples include the advent of big web-based corpora, such as the 20.1-billion word NOW corpus (Davies 2016) or the multilingual TenTen corpora available in Sketch Engine (Kilgariff et al. 2014); large corpora of social media data such as Twitter and YouTube (Bonilla et al. 2024; Coats 2023; Grieve et al. 2018; Morin and Coats 2023; Morin and Grieve 2024); and corpora spanning an increasingly wide typological scope of regional varieties and registers (Dunn 2020).

This radical shift in linguistic methodology has brought with it the unique asset of providing enough synchronic data to collect quantitatively representative samples of language phenomena from vast areas around the globe. However, this evolution has also come with an outstanding practical problem, which represents the core issue considered in this paper. When faced with such large quantities of data, it is often necessary to manually clean overwhelming quantities of tokens. As an example of this issue, let us introduce the case study at the heart of this paper, which involves the verb *consider* in English. *Consider* is a verb that can be found in a number of larger grammatical constructions, one of which is an evaluative use introducing an evaluation or judgement of something or someone (Jacques 2022).[1] Although there is little previous research on the evaluative *consider* construction in English, it appears to display variation in its morphosyntactic realizations, with at least three options, exemplified in (1) from the 2012 blog section of the Corpus of Contemporary American English (COCA; Davies 2008–). The first is a bare realization of the verb directly followed by the subject complement, as in (1a). The second is a realization with the preposition *as* making the object complement indirect, as in (1b). The third is a realization with an infinitival clause *to be* introducing a predicative complement, as in (1c).

**\*Corresponding author: Matti Marttinen Larsson,** University of Gothenburg, Gothenburg, Sweden,
E-mail: matti.marttinen.larsson@gu.se. https://orcid.org/0000-0002-6224-7872
**Cameron Morin,** Université Paris Cité, Paris, France. https://orcid.org/0000-0001-7079-449X

(1)  a.  *Cain is **considered** a master of American hard boiled crime fiction (…)*
     b.  *By the way, I also **consider** Zelda II **as** one of my favorites (…)*
     c.  *Since he **considers** himself **to be** a man of God, one would think he would know that it is clearly stated in the Bible that a Christian is NEVER to question the faith of another.*

The relationship between these three variants is unclear and has not been explained in previous research. For example, one relevant question that surfaces is whether the competing constructions are semantically, pragmatically, or socially distinct (cf. Leclercq and Morin 2023). It has also been suggested by dictionaries of English usage such as Merriam-Webster that "the *consider* + *as* construction is becoming less and less common", and that the bare form appears to be the most idiomatic variant (Britannica Dictionary n.d.). We could therefore pursue a diachronic line of enquiry and analyse potential language change in historical corpora.

The questions raised above would generally call for a corpus-based study. Crucially, however, the researcher may well be hard-pressed to conduct such a study when attempting to collect relevant data on the *consider* construction from large corpora. Indeed, searching the lemmatized form CONSIDER in the EnTenTen21 corpus using Sketch Engine, we find no fewer than 18 million tokens, which far exceeds any human capacity to sort through. A common solution to this quantitative issue is to either extract a random sample from the corpora and proceed with the annotation of a more manageable amount of data, or to annotate data automatically using natural language processing (NLP) techniques (e.g., spaCy; Honnibal and Montani 2017). However, in the case of the *consider* construction, the problem remains that uses of the verb as an evaluative construction are far from exhausting all possible instances of *consider* hits. Indeed, the evaluative construction coexists with another prominent use of *consider* as a cognition verb,[2] as shown in examples (2a)–(2c) (again, taken from the blog section of COCA; Davies 2008–). In these examples, *consider* means 'take into account', 'believe/think', and 'contemplate', respectively.

(2)  a.  *Please **consider** all classes when punching number in to spread-sheet and designing new items.*
     b.  *Perhaps Turbine just **considered** that his work was done as all the systems were in place (…)*
     c.  *I hope he **considers** running in 2016!!*

To get a sense of how difficult it is to find the evaluative construction in an unsorted collection of data points, we extracted a random sample of 200 tokens of CONSIDER from the EnTenTen2021 corpus (see supplementary material). Following a manual clean of this sample, out of these 200 tokens, only 11.5 % were found to be true positives, that is, actual data of interest for the research questions at hand.

As we hope to have made clear by this point, a dedicated corpus-based study of a grammatical construction such as *consider* poses a significant methodological challenge, and this is likely to hold for many other types of constructions in English and other languages. It is indeed often a time-consuming, fatigue-generating, and hence cognitively inefficient (Brazaitis and Status 2023) preliminary task to sift through large samples of data in search of the construction of interest. This challenge can be compounded by the amount of variation in patterns

---

**1** Jacques (2022: 178–179) refers to these as "estimative" and conflates multiclausal [*consider* + complementizer *that* + clause] constructions with the evaluative monoclausal [*consider* X *as* Y]. Our analysis differs in this regard. First, it seems to us that the two types differ semantically and pragmatically: whereas monoclausal [*consider* X *as* Y] directly introduces an evaluation, classification, or judgement about something or someone, multiclausal [*consider* + complementizer *that* + clause] involves deliberation and is more indirect. This is why we consider the latter a use of *consider* as a cognition verb rather than evaluative. Second, they are structurally highly distinct. We do therefore not consider multiclausal *consider* to form part of the envelope of variation.

**2** See, for instance, the inclusion of the verb *consider* in a glossary of cognitive verbs by the Queensland Curriculum and Assessment Authority (2018).

co-occurring with the construction. For instance, any corpus-based inquiry of evaluative *consider* would need to include its potential interaction with passive constructions, in addition to the three morphosyntactic realizations presented above. This would give us no less than six sets of requests to handle simultaneously.

Taking the example of *consider*, in this paper we showcase a methodological solution to these problems which makes use of large language models (LLMs). In particular, we show that LLMs can be trained to more quickly, more efficiently, and perhaps even more reliably deal with cleaning and annotation tasks for studies of grammatical constructions and grammatical variation. Notably, we outline a sequence of steps which can be taken to automatically annotate large quantities of evaluative *consider* constructions, based on the key processes of prompt engineering, training, and evaluation. Our case study makes use of the LLM Claude 3.5 Sonnet (Anthropic 2024a) and data from the NOW corpus and Sketch Engine's EnTenTen21 corpus. Our ultimate goal in this paper is twofold:

(1) To build a replicable methodological pipeline that can be repurposed for the automatic annotation of a wide array of grammatical constructions in English and other languages

(2) To substantially improve the quality of annotation and "quality of life" of future corpus studies of grammatical constructions, using LLMs as "copilots for linguists" (Torrent et al. 2024)

The rest of the paper is structured as follows. In Section 2, we briefly review existing research on the use of LLMs as tools for linguistic studies. In Section 3, we take the case of *consider* constructions to delineate a replicable method for the automatic, supervised annotation of grammatical data, broken down into three main steps: prompt engineering (pre)training, and evaluation. Section 4 concludes on the generalizability of our findings and some future directions for LLM-assisted research in grammatical variation and change, in addition to noting some important caveats for an appropriate and responsible use of the method.

## 2 Large language models

The use of LLMs in linguistics represents a booming research domain with a wide array of applications and questions. To mention but a couple of examples, LLMs have sparked important theoretical discussions of their potential similarities with cognitive models of human linguistic processing, knowledge, and use (e.g., Cuskley et al. 2024; Goldberg 2024; Leivada et al. 2024; Piantadosi 2023; Weissweiler et al. 2023); and they have been applied for modelling processes of sociolinguistic variation (e.g., Grieve et al. 2025; Hekkel et al. 2024; Lilli 2023; Massaro and Samo 2023). In this section, we focus specifically on studies that have investigated how LLMs can be leveraged as "copilots for linguists" (Torrent et al. 2024), as long as they are handled with caution and carefully supervised (Denning et al. 2024; Ollion et al. 2023; see also Section 4).

One of the most popular LLMs for automatic data annotation in linguistics is ChatGPT (e.g., GPT-4; OpenAI 2024). Pioneering studies have suggested that ChatGPT can be an effective and reliable tool for a range of different topics, such as sentiment analysis (Belal et al. 2023), the evaluation of corpus annotation schemes in treebanks (Akkurt et al. 2024), and the annotation of pragmatic and discourse related phenomena in corpora (Yu et al. 2024). In this paper, we use the LLM Claude 3.5 Sonnet (Anthropic 2024a; henceforth "Claude"), which has seen a rise in usage for NLP studies, including in comparison with ChatGPT (e.g., Caruccio et al. 2024; Kholodna et al. 2024).

NLP-oriented LLMs such as spaCy (Honnibal and Montani 2017) already include specific features to assist computational linguists in the annotation of various types of data. However, there exist no user-friendly methodological pipelines for grammatical annotation that are accessible to the linguistic community at large. This is precisely the gap that we seek to fill with this paper: not only do we seek to introduce an AI-assisted method for the annotation of specific grammatical constructions and their variation patterns for the first time, but we also seek to introduce a method that is accessible to a maximally wide usership, given that it does not require advanced skills in coding, unlike with computational linguistic methods such as spaCy. Indeed, Claude, similarly to ChatGPT, is a conversational AI that is accessible through a chat interface, as opposed to spaCy; the key skills to master for the user, then, are training and prompting. Prompt engineering consists in designing and refining text inputs to an LLM for it to learn from mistakes and "reason better" (OpenAI and Ekin 2023; Shengnan et al. 2024). For the specific task of automatic linguistic annotation, prompt engineering is essential, as it enables the researcher to supervise and improve the accuracy of the LLM in completing the task. We outline a methodological pipeline for this task in the next section.

# 3 Automating grammatical annotation: an iterative process applied to *consider*

In this study, we built an iterative process in Anthropic's Claude[3] (used in November 2024) to design a pipeline for the automatic annotation of grammatical variation in evaluative constructions with the verb *consider*. The iterative process ties in with the LLM's ability to "learn" from mistakes (Wei et al. 2022): by working through examples of corpus data on a case-by-case basis in the chat interface during the training phase, Claude gains an increasingly better understanding of the classification criteria and refines the instructions through training. This training is contained within the realms of the project, and it does not extend further; in other words, the capabilities acquired and applied by Claude are constrained to a particular conversation and cannot be accessed through another chat outside of that chat (Anthropic 2024b).

In what follows, we outline the main steps involved in an automatic annotation pipeline in Claude. We also discuss different strategies that can be used to enhance Claude's performance. These strategies are all based on Anthropic's guide *Build with Claude* (Anthropic n.d.), which is the main source for the rest of this section.

## 3.1 Prompting

The first step of the pipeline consists in prompt engineering, that is, the formulation of input prompts that help elicit the desired response from Claude (Anthropic n.d.; White et al. 2023). The full prompt designed for the present case study can be found in the supplementary material. LLMs such as Claude are highly sensitive to the formulation of prompts; therefore, we propose four design principles:

(1) Make the prompt clear, specific, and contextualized: This involves explaining (a) what type and purpose of the task Claude should perform (e.g., classification), as well as what constitutes a successful task completion (e.g., the desired accuracy level); (b) specific instructions in sequential steps, such as bullet points or numbered lists; and (c) telling Claude in clear and explicit terms what type of classification it should return (e.g., LLM-based grading) and how the outcome will be evaluated.

(2) Include examples: This involves providing a small number of on-point examples in the prompts, which allows Claude to better understand the type of classifications that are involved in the task prior to (pre)training.

(3) Include XML tags: This involves helping Claude process the prompts more accurately. Some useful tags include <example> </example> and <examples> </examples> to wrap examples; <instructions> </instructions> to wrap specific instructions that Claude should consider; and <thinking> </thinking> to wrap step-by-step Chain of Thought prompting (see below).

(4) Tell Claude to think: When Claude is explicitly instructed to reason about its classification and to "think step by step", it breaks down the classification process into a step-by-step process. This is referred to as Chain of Thought prompting, which significantly increases Claude's performance in dealing with complex and demanding tasks.

We now turn to the iterative process underlying Claude's data classification.

---

[3] Claude is available for free, but a professional plan can also be purchased to access the LLM at higher rate limits.

## 3.2 Pretraining, iterative training, validation, and evaluation

Figure 1 illustrates the pipeline that we utilized in the present study. A key feature of our approach is that LLMs such as Claude receive substantial benefits in performance through iterative refinement. This involves the synergy between training and validation, which can be repeated until a desirable accuracy is reached. The second part of the pipeline is thus iterative, consisting of a back-and-forth training process.

First, we fed Claude with a dataset of approximately 500 pre-classified corpus sentences of *consider* that contained a binary classification (evaluative vs. non-evaluative, extracted from the NOW corpus). These observations needed to contain data that accurately reflects the target, including edge cases (e.g., irrelevant and ambiguous sentences; Anthropic n.d.). At this point, Claude was instructed to think[4] about how the data had been classified, and to think step by step to arrive at a conclusion regarding whether it would have classified the data in the same manner, considering the provided instructions. Claude was also encouraged to ask questions and make comments.

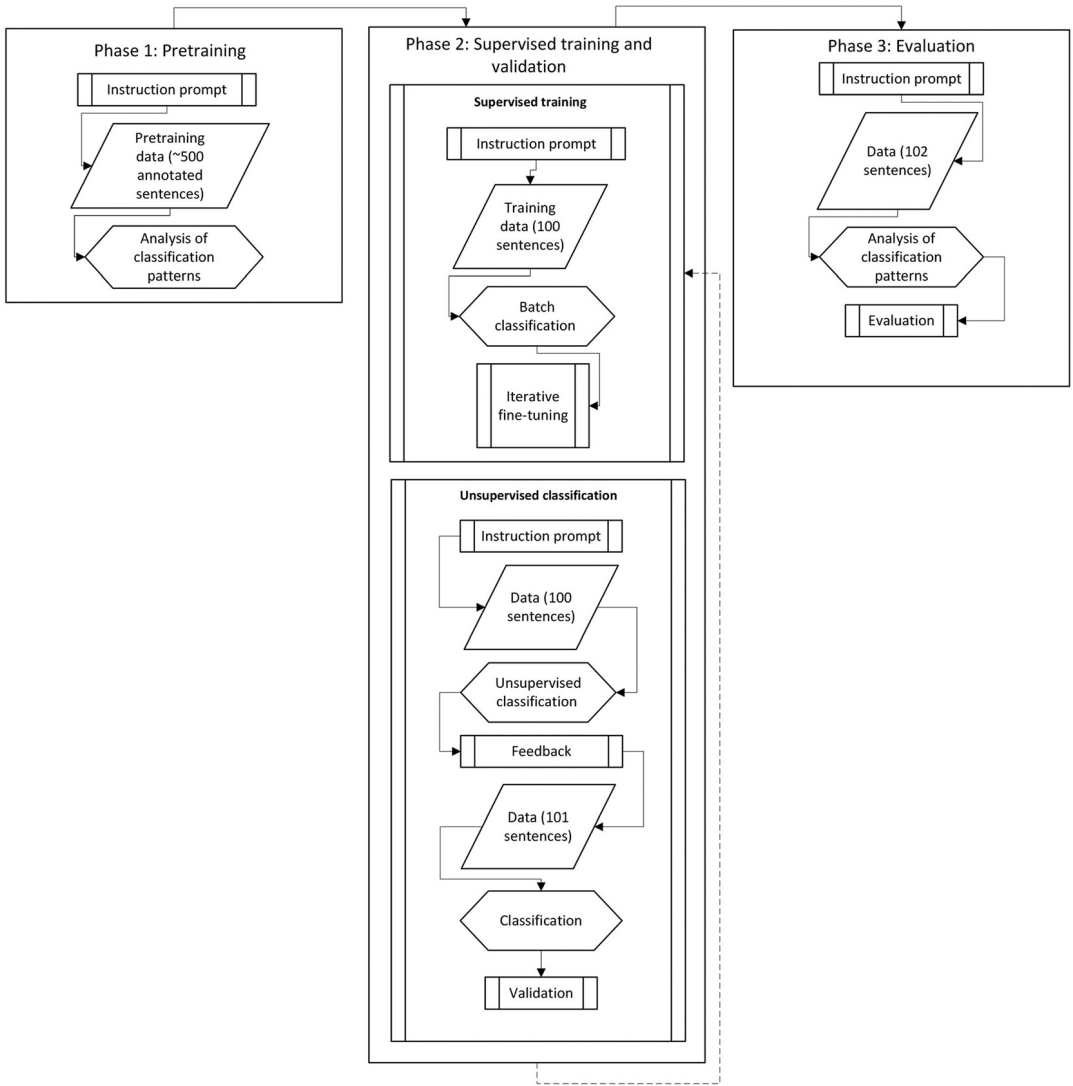

**Figure 1:** Methodological pipeline for LLM-assisted annotation of linguistic data.

---

4 The use of cognition-related terms such as *understand, think*, and *reason* in reference to LLMs is intended as a convenient shorthand rather than a claim about their cognitive capacities and should be interpreted metaphorically rather than as an assertion of human-like cognition (cf. Bender and Koller 2020; Hazra et al. 2024; Kambhampati 2024).

**Table 1:** Evaluation of Claude's binary classification of (non-)evaluative *consider*.

| | |
|---|---|
| Accuracy | 93 % (95/102) |
| Precision | 88.1 % |
| Recall | 94.9 % |
| F1 score | 91.4 % |
| Matthews correlation coefficient | 0.86 |

The second step involved two sub-processes: "supervised training" and "unsupervised classification" with validation. Claude was first instructed to provide a classification of a training set with 100 *consider* sentences. Working in batches of 20–25 sentences, Claude was encouraged to think step by step and include <thinking> </thinking> and <answer> </answer> for each classification. For each batch, Claude received corrective feedback from the user, informing Claude of what classifications were wrong and why they were wrong. Each round of corrective feedback constituted input that Claude used as its basis for further refinement of classification criteria, leading to an increasingly higher accuracy.

Following the supervised training, the next step involved the unsupervised classification of 100 unseen sentences. After having evaluated the unsupervised classifications (which had an accuracy of 67 %), Claude received corrective feedback (e.g., level of accuracy, examples of misclassifications, new instructions clarifying any misunderstandings). This iterative training and validation cycle was repeated until Claude achieved the desired level of accuracy, with each round of classifications always providing unseen data. Importantly, throughout the conversations, opportunities emerged for Claude to propose new guidelines of analysis, decision criteria, comments, and questions, which could constitute valuable clues to determine whether Claude had understood the assignment correctly. The user can correct Claude at any point of the interaction, allowing Claude to update its knowledge within the scope of the conversation.

Finally, for the case study at hand, following pretraining, supervised training, and unsupervised training, Claude was instructed to classify this dataset while taking into account the refinements that it had gained from the previous rounds and it was presented with 101 more sentences for validation. For this validation dataset, Claude reached an accuracy of 93 % (94/101). The final evaluation was made on a dataset of 102 *consider* sentences that were annotated blindly to Claude's output. The evaluation metrics, which indicate a strong performance, are presented in Table 1.[5] The process from pretraining to evaluation took approximately 60 min to complete, illustrating the substantial time-saving value of the methodological pipeline.

# 4 Conclusion, caveats, and outlook

In this paper, we presented a methodological pipeline leveraging a large language model for the automatic annotation of grammatical data, based on the case study of the evaluative *consider* construction in English. By following the key steps of prompt engineering, iterative training, and evaluation, we succeeded in building a model that was successful at automatically annotating unseen samples of the construction with a high rate of accuracy. In the specific case of evaluative *consider*, we are now able to address at scale the research questions sketched in Section 1: are the three evaluative variants equally common in present-day English, and if so, are they equivalent in meaning? Over time, do we find shifts in frequencies of these variants, and could we be dealing with

---

**5** On 24 February 2025, a new Claude Sonnet model was released (Claude 3.7 Sonnet). To assess the direct replicability of our training, we utilized the transcript and data used in Phases 1 and 2 (see Figure 1) as input in Claude 3.7 Sonnet, ensuring that it got the exact same input and fine-tuning as the model reported on in this paper. This was done on a different computer in another country, on another user account (this time, CM's account rather than MML's). We subsequently fed Claude 3.7 Sonnet with the same evaluation dataset that we used to evaluate 3.5 (reported on in this paper). This rendered highly similar evaluation scores: accuracy 92 %; precision 94 %; recall 85 %; F1 score 89 %; Matthews correlation coefficient 0.83.

a case of grammaticalization towards reduction (i.e., shortening, *consider as > consider Ø*; Britannica Dictionary n.d.; Levshina 2022; Marttinen Larsson 2024) or towards enhancement (i.e., lengthening, *consider as/Ø > consider to be*; Levshina 2022; Marttinen Larsson in press)?

From a more general standpoint, in this paper we put forward this annotation method as a replicable one, which linguists can use for the analysis of other grammatical constructions, in English and in other languages. The individual parameters, instructions, and evaluation criteria are of course bound to vary depending on the specific construction under study. In addition, it is reasonable to assume that some languages (especially English) will be easier to study than others depending on the amount of available training data in LLMs. Overall, however, we believe that this pipeline can be used as a tool for any linguist interested in analysing grammatical constructions with large amounts of corpus data and the assistance of an AI copilot.

This paper has focused on a number of clear advantages presented by the method for future research, but there are also a number of caveats to be made.

First, the method presented in this paper offers a new approach to corpus annotation, but it is not intended to replace more traditional processes involving one or multiple annotators, particularly for tasks involving semantic or pragmatic ambiguity in linguistic constructions. In this study, we focused on a relatively clear-cut formal case, where inter-annotator disagreement is likely to be minimal. However, for more complex semantic classifications, it remains to be explored how LLMs such as Claude perform (for early work in this direction, see Yu et al. 2024).

Second, this paper focused on the use of one specific LLM, namely Claude, for the elaboration and the execution of the methodological pipeline which we put forward. However, the choice of LLM for prompt-based annotation may be a crucial factor, as different models have distinct training data and optimization strategies that can influence their performance. While some models, such as ChatGPT, undergo reinforcement learning to refine their outputs in ways that obscure the original distribution, others, like LLaMA, avoid such adjustments but still have opaque training data. Transparent models such as Pythia offer the advantage of allowing researchers to examine potential training biases directly. In this study, we focused on Claude because it provides a powerful and accessible chat-based interface that does not require programming expertise, making it a practical tool for linguists seeking to repurpose this method for their own research. Moreover, the systematic prompt engineering and training steps in our approach help mitigate potential biases in Claude's training data. Future research could explore how different models, particularly those with transparent training data, compare in their ability to perform annotation tasks reliably.

Third, another concern pertains to the legal aspects of training LLMs on corpus data. Given that some LLMs learn from the input they receive, utilizing input data that contains sensitive information or that is in some way access-restricted could be problematic. It therefore seems imperative that the type of data that researchers feed into an LLM is duly considered before any LLM-assisted annotation task is initiated. Similarly, the type of LLM utilized in such tasks is also important from a legal standpoint, as some explicitly state that they do not use the received input or generated outputs to train their model. Claude's policy of not utilizing user interactions as input represents an advantage in this respect.

Fourth, this approach raises new questions worth exploring. Since our study proposes a pipeline for annotating variable phenomena, it is crucial to consider that many such phenomena reflect ongoing language change. The incoming variant may be marginal, causing a construction to exhibit both central and peripheral uses, which could make it harder for the LLM to correctly identify peripheral cases. This issue relates to two aspects of the data: (i) the distribution of "conventional" versus "incipient" variants across contexts; and (ii) the distinction between "easier" and more complex occurrences (edge cases; cf. Section 3). One such edge case arises when intervening material separates *consider* from its complement clause, complicating identification (cf. Gibson 1998). However, the boundary between centre and edge is often unclear, making it difficult to define a stratified sample for training and evaluation. Instead, we include a representative corpus sample in training, ensuring the LLM learns the envelope of variation. If edge cases are pervasive, they will be reflected in the training data. Regarding variant distribution, careful design of training data is necessary. If prior research indicates ongoing change, and the study involves incipient change or diachronic analysis, future work could mitigate this issue by using stratified training data – for example, data from different time periods or a balanced distribution across variants. In addition, evaluation metrics are critical. Given the likelihood of imbalance in diachronic or variationist studies, it is essential to consider all dimensions of the confusion matrix, not just correct classifications. We use the Matthews correlation coefficient for this purpose (Table 1; see also Chicco and Jurman 2020; Marttinen Larsson 2023).[6]

Finally, a similar question concerns LLMs' ability to classify language change in meaning rather than form: can they accurately identify forms exhibiting layering due to ongoing change? While beyond this study's scope, recent research has explored this with other LLMs. Bonilla et al. (2025) analyse how Spanish BERT (BETO) interprets *literalmente* 'literally', which is undergoing grammaticalization. Over time, this adverb has shifted from involving word-by-word denotation to developing increasingly expressive meanings (viz. intensification of hyperbole, and truth-value emphasis). Using local interpretable model-agnostic explanations and Universal Dependencies, Bonilla et al. (2025) examine BETO's classification of these three meanings and find that, while BETO struggles with the most grammaticalized and more "pragmatic" uses, it still performs well in classifying these non-etymological meanings. Moreover, it relies on cues symptomatic of grammaticalization, such as increased syntactic flexibility and broader modification of part-of-speech tags. This suggests that BETO quite effectively identifies semantic layering despite formal stability. Future research should examine how other LLMs, such as Claude, perform on similar tasks.

With these considerations in mind, we feel that the following guidelines could serve as a basis for future responsible and carefully calibrated applications of the LLM-assisted pipeline outlined in this study:

– The training data should accurately reflect the data that the LLM will classify. For instance, it would not be appropriate to use training data from spoken informal corpora for a classification task that will be performed on written formal data, unless the empirical issue at hand is to assess performance differences that may be attributable to inter-register variability. The same concern applies to time periods, dialectal varieties, and so on.
– LLMs should not be used blindly. They should only come into play after substantial preparatory work on the grammatical construction, including a comprehensive understanding of the envelope of variation and its possible edge cases. Only following these steps can the LLM be responsibly used for annotation tasks.
– The brittle nature of prompt engineering should be kept in mind. LLMs are highly sensitive to wording. This paper hopes to inspire careful elaboration during prompt engineering, but caution is required.
– While more research is needed to determine LLMs' classification performance on a greater variety of linguistic phenomena, it seems that, as it currently stands, prompt-based LLM classification using chatbot interfaces (e.g., Claude) is perhaps most sensibly used for more straightforward classification tasks (such as classifying differing surface realizations of *consider*), whereas other more semantically and pragmatically complex expressions require due caution and perhaps even more rigorous training and evaluation.

Some remaining questions that were beyond the scope of this paper constitute important future directions. For one, the present paper focused on illustrating the use of LLMs for extracting relevant data among large quantities of corpus occurrences. Future research should ascertain the performance of Claude in annotating predictor variables of potential relevance in subsequent analyses. Moreover, designing a comparative study of the performance of Claude with that of other well-known, user-friendly LLMs, such as ChatGPT, is a desirable venture for the optimization of the method introduced here. Furthermore, a more direct comparison between the performance of these LLMs and humans on comparable samples of corpus data may highlight their value for prolonged annotation sessions, which will necessarily result in accumulated fatigue for the annotator. For the time being, however, we hope to have shown that the methodology sketched here already reaches satisfactory levels of accuracy, and represents sufficiently valuable cost efficiencies in time, energy, and data quality consistency, to be seriously considered in future linguistic studies of this type.

---

**6** Matthews correlation coefficient is an evaluation metric particularly apt for dealing with unbalanced data, because it not only takes into account the percentage of true predictions (accuracy) nor the harmonic mean between precision and recall (F1), but it also takes into account negative elements (Chicco and Jurman 2020).

# Supplementary material

The pretraining data, annotated datasets, and prompt histories for the case study of the paper are available on an OSF database at https://doi.org/10.17605/OSF.IO/TYJZ6.

**Acknowledgements:** We would like to thank Jesper Håkansson for his sustained support and advice on training and prompt engineering using Claude 3.5 Sonnet.